\documentclass[letterpaper, 10 pt, conference]{ieeeconf}  

\IEEEoverridecommandlockouts                              

\usepackage{amsmath} 
\usepackage{amssymb}  

\usepackage{cite}
\usepackage[pdftex]{graphicx}
\usepackage{algorithm}
\usepackage{algorithmicx}
\usepackage{algpseudocode}
\algrenewcommand\algorithmicrequire{\textbf{Input:}}
\algrenewcommand\algorithmicensure{\textbf{Output:}}
\usepackage{array}
\usepackage{multirow}
\usepackage[font=small]{caption}
\usepackage{subcaption}

\usepackage{tikz}
\usepackage{pgfplots}
\usepackage{xcolor}
\pgfplotsset{compat=1.17}
\usetikzlibrary{positioning}
\usetikzlibrary{decorations.pathreplacing}
\usetikzlibrary{calc}
\usetikzlibrary{positioning,shapes,shadows,arrows}

\pgfplotsset{
every axis/.append style={
  axis line style={->}, 
  legend style={font=\scriptsize},
  label style={font=\scriptsize},
  title style={font=\scriptsize},
  tick label style={font=\scriptsize},
  }
}

\makeatletter
\let\NAT@parse\undefined
\makeatother
\usepackage{hyperref}

\title{\LARGE \bf Rearrangement-Based Manipulation via Kinodynamic Planning and Dynamic Planning Horizons}

\author{Kejia Ren, Lydia E. Kavraki and Kaiyu Hang
\thanks{The authors are with the Department of Computer Science, Rice University, Houston, TX 77005, USA. \{\tt\small kr43, kavraki, kaiyu.hang\}@rice.edu.}%
\thanks{In this work, KR and KH are supported by NSF CMMI-2133110 and Rice University Funds, and LEK is supported in part by NSF 2008720.}%
}

\begin{document}

\maketitle
\thispagestyle{empty}
\pagestyle{empty}

\begin{abstract}

Robot manipulation in cluttered environments often requires complex and sequential rearrangement of multiple objects in order to achieve the desired reconfiguration of the target objects. Due to the sophisticated physical interactions involved in such scenarios, rearrangement-based manipulation is still limited to a small range of tasks and is especially vulnerable to physical uncertainties and perception noise. This paper presents a planning framework that leverages the efficiency of sampling-based planning approaches, and closes the manipulation loop by dynamically controlling the planning horizon. Our approach interleaves planning and execution to progressively approach the manipulation goal while correcting any errors or path deviations along the process. Meanwhile, our framework allows the definition of manipulation goals without requiring explicit goal configurations, enabling the robot to flexibly interact with all objects to facilitate the manipulation of the target ones. With extensive experiments both in simulation and on a real robot, we evaluate our framework on three manipulation tasks in cluttered environments: grasping, relocating, and sorting. In comparison with two baseline approaches, we show that our framework can significantly improve planning efficiency, robustness against physical uncertainties, and task success rate under limited time budgets.

\end{abstract}

\section{Introduction}
\label{intro}

Research in robotic manipulation has investigated how to reconfigure objects in different task scenarios and robot-object-environment formulations, such as grasping, pick-and-place, in-hand manipulation, dual-arm manipulation, objects sorting, and placement \cite{billard2019trends, Bohg14, hang2021manipulation}. Traditionally, most manipulation tasks have been studied as standalone problems without considering the physical interactions with any other objects. For example, grasping is often modeled as a static process where a hand needs to reach a stable grasp on the target object without touching anything else. Although such isolated formulations can simplify the problem and are in many cases sufficient, e.g., in-hand manipulation only concerns a hand and the grasped object, they are inherently oversimplified and not applicable to many real-world settings, since the target objects are not always located in free spaces. Importantly, even if we can sequentially manipulate one object at a time to achieve certain goals, concurrent manipulation of multiple objects has proven a much more efficient strategy for various tasks \cite{lynch1999dynamic}, especially when multiple objects need to be relocated relatively to each other.

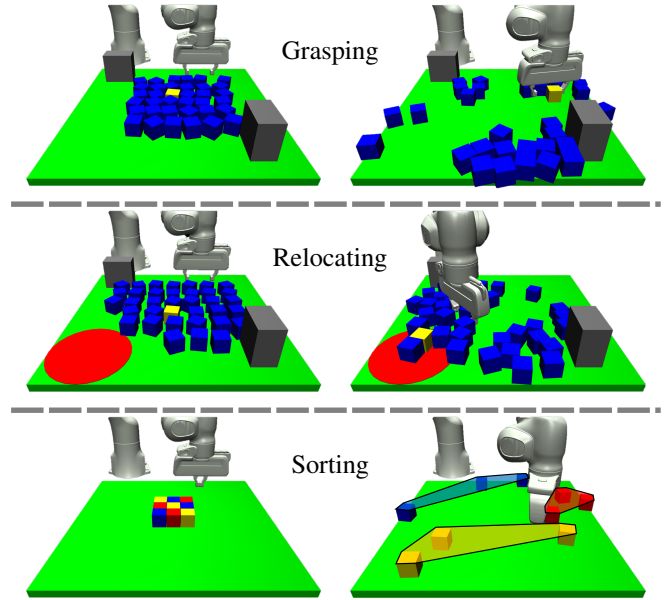
\begin{figure}[t]
\centering
    \begin{tikzpicture}
        \node[anchor=south west,inner sep=0] at (0,0){\includegraphics[width=\linewidth]{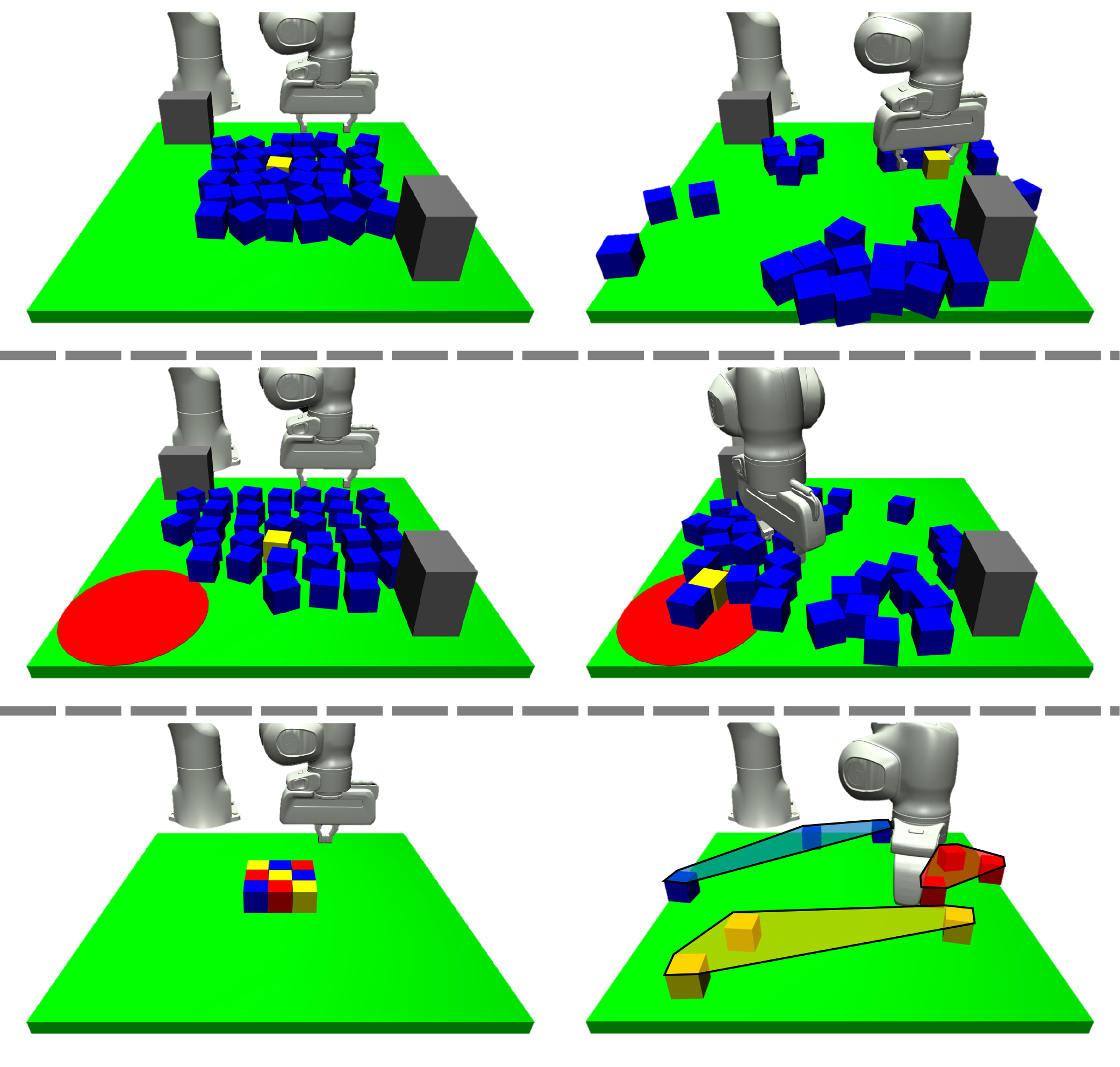}};
        
        \node[align=center] at (4.25, 7.5) {Grasping};
        \node[align=center] at (4.25, 4.75) {Relocating};
        \node[align=center] at (4.25, 2) {Sorting};
    \end{tikzpicture}
    \caption{Example rearrangement-based planning tasks. The left column shows the start states, and the right column shows the goal states achieved by our proposed planner. \emph{Grasping}: The robot approaches a grasp for the target object (yellow) surrounded by other movable objects (blue) while avoiding collisions with the obstacles (gray); \emph{Relocating}: The robot pushes the target object (yellow) to a goal region (red); \emph{Sorting}: The robot rearranges to separate objects of different classes into clusters.}
\label{fig:first}
\vspace{-0.7cm}
\end{figure}

As such, rearrangement-based manipulation, defined as a class of problems requiring a robot to concurrently manipulate multiple objects to achieve the task goal, has been broadly studied \cite{ruggiero2018nonprehensile, woodruff2017planning}. A few example tasks are illustrated in Fig.~\ref{fig:first}. Nevertheless, it still remains a difficult class of problems because: 1) it has proven to be NP-hard \cite{wilfong1991motion}; and 2) as the problem space is composed of the robot configuration, robot control, and the configurations of all objects, the dimensionality of such problems is much higher than most other manipulation problems, rendering them computationally very challenging. Such challenges can be intuitively seen in the example grasping task, where the robot needs to rearrange the surrounding objects so that the gripper can reach a stable pre-grasp pose. A major difficulty of this task is that, while the surrounding objects are being rearranged, the target object is simultaneously moved by multiple object-object interactions.


Moreover, almost all existing nonprehensile rearrangement planners rely on either hand-crafted physics models or simulators to model the system state transitions. Such methods usually assume very precise models, e.g., the geometries and physical parameters of objects to be available during the planning time, which is in general infeasible. This fact brings up a significant problem: the inherent discrepancies between the system models and the real world will accumulate errors throughout the entire execution of the planned actions. Thus, even if a planner has successfully generated a plan, the real execution will likely deviate the manipulation from the desired path and cause task failures. In other cases, if the initial perception of the system's state is not reliable \cite{butepage2019visual}, or if the system's state has been changed during execution due to human interruptions or environment changes, most open-loop approaches will merely continue executing without the capability to actively correct the errors. 

In this work, inspired by dynamic window-based approaches for robot navigation in partially observable environments \cite{Bekris07}, we propose a kinodynamic planning framework for rearrangement-based manipulation with dynamic planning horizons.
In brief, our approach is able to:
\begin{enumerate}
 \item monitor the planning progress and dynamically determine the planning horizons, and direct the planning into more task-relevant subspaces to significantly improve the planning efficiency;
 \item react to physical and perception uncertainties online, and work with imperfect system models, e.g., inaccurate object geometries, to progressively generate and execute rearrangement actions while correcting observed errors;
 \item address various rearrangement problems, with and without explicitly defined goal configurations, to allow the robot to flexibly interact with all objects to facilitate the manipulation of the target objects.
\end{enumerate}

\section{Related Work}
\label{sec:relatedwork}


\emph{Motion Planning:} In physical interaction-based motion planning problems where the robot-object-environment configurations are constantly changed, kinodynamic planning has been investigated to jointly model the robot configuration, the robot control, and the system transitions \cite{lavalle2001randomized}. Among other approaches, sampling-based kinodynamic planning algorithms have been widely employed due to its great efficiency and generalizability. As compared with kinodynamic problems in collision-free environments \cite{lau2009kinodynamic}, however, kinodynamic manipulation planning is much more complex due to the challenges of the dramatically increased problem dimensionality, highly nonlinear physics, and uncertainties in perception. Rearrangement-based manipulation is in particular a difficult set of problems for kinodynamic planning. In addition to the aforementioned challenges, as rearrangement is often about the relative reconfiguration between objects, it is infeasible for a planner to always explicitly define goal configurations. In this work, inspired by the work of robot navigation planning in partially observable or dynamically changing spaces \cite{Bekris07, hsu2002randomized}, we introduce progress control into kinodynamic manipulation planning. By dynamically adapting the planning horizon, our method is able to progressively plan the manipulation motions with significantly improved efficiency, and can handle problems without explicitly-defined goal configurations.

\emph{Planning-based Rearrangement:} Kinodynamic RRT-based planning algorithms have shown promising potentials in rearrangement tasks. Using a problem-specific contact model under quasi-static assumptions, \cite{king2015nonprehensile} analytically plans a diverse set of pushing motions but prohibits object-object interactions. Based on an efficient physics simulator, multi-object interactions are enabled, and dynamic motions of objects, e.g., rolling, can be incorporated \cite{haustein2015kinodynamic}. Additionally by modeling the uncertainties in physics \cite{moll2017randomized}, or optimizing a continuous motion trajectory online \cite{agboh2018real}, grasping in cluttered environments has been achieved by locally rearranging the occluding objects. Further, rearrangement tasks with relative goals, e.g., sorting, have been addressed by learning-based Monte Carlo Tree Search \cite{song2020multi}, and iterative local search to concurrently manipulate a large amount of objects \cite{huang19}. However, the existing approaches either are not able to address the physical uncertainties during execution, or require very complex modeling of physics and sophisticated problem-specific heuristics, making them difficult to be easily generalized to various rearrangement-based manipulation problems. In contrast, based on any physical simulators, even without precise physical models, our proposed framework is able to react to physical uncertainties online, and generalize to complex tasks based on simple heuristics.

\emph{Learning-based Rearrangement:} Recently, data-driven rearrangement planning has been extensively studied to tackle various tasks. In end-to-end settings, pushing-based relocation \cite{yuan2019end}, multi-object rearrangement and singulation \cite{eitel2020learning, hermans12}, rearrangement-based grasping \cite{laskey2016robot}, etc., have been formulated as policy-learning problems to reactively generate robot actions online. Although such approaches have greatly simplified the system pipeline and allow for direct action generation purely based on the input images, as a common challenge, they in general require a large amount of training data for specific tasks, while the learned models are difficult to be transferred to achieve different tasks \cite{Kaelbling915}.

\section{Problem Statement}
\label{sec:problem}

We formulate rearrangement-based manipulation planning as a kinodynamic motion planning problem. Given a bounded workspace $\mathcal{W} \subset SE(2)$, containing a robot manipulator, $N$ movable objects to manipulate, and a set $\mathcal{O}$ of static obstacles, we aim to find a sequence of robot motions, called a motion plan, such that the environment will be rearranged to reach a state satisfying the goal criterion.

\subsection{Terminology}

\subsubsection{State Space} 

Formally, we denote the robot state as $q^R \in \mathcal{Q}^R \subset \mathbb{R}^r$, where $\mathcal{Q}^R$ is the robot configuration space and $r \in \mathbb{R}$ is the robot's degree of freedom. Let the state of a movable object be $q^i \in \mathcal{Q}^i \subset \mathcal{W}$, where $\mathcal{Q}^i \subset SE(2)$, and $i \in \{1, ..., N\}$. The state space of the planning problem is then defined by the Cartesian product $\mathcal{Q} = \mathcal{Q}^R \times \mathcal{Q}^1 \times . . . \times \mathcal{Q}^N$, and a system state $q \in \mathcal{Q}$ is denoted by a tuple $q = (q^R,  q^1, ..., q^N)$. A state $q$ is valid only when the robot does not collide with itself or any static obstacle in $\mathcal{O}$, and all the movable objects are inside the workspace $\mathcal{W}$. All the valid states compose the valid state space $\mathcal{Q}^{valid} \subseteq \mathcal{Q}$. Note that, $\mathcal{Q}^{valid}$ is different from the $\mathcal{C}^{free}$ space in traditional motion planning problems, as the contacts between movable objects and static obstacles, as well as between any pair of movable objects or the robot are allowed for a valid state.

\subsubsection{Control Space and Transition Function} 

The control space $\mathcal{U} \subset \mathbb{R}^r$ is a sampleable continuous space consisting of all controls the robot is allowed to perform. 
We denote by a transition function, $\Gamma: \mathcal{Q}^{valid} \times \mathcal{U} \mapsto \mathcal{Q}$, to represent the physics laws of the real world, which maps a state $q_t \in \mathcal{Q}^{valid}$ and a control action $u_t \in \mathcal{U}$ at time $t$ to the state outcome at the next step $q_{t+1} \in \mathcal{Q}$.

\subsubsection{Goal Criterion} 
We define the goal criterion as a function $g: \mathcal{Q}^{valid} \mapsto \{0, 1\}$ specified by the manipulation task. Therefore, the goal region of the planning problem is defined by a set $\mathcal{Q}_G = \{q \in \mathcal{Q}^{valid} \; | \; g(q) = 1\}$ of all states that satisfy the goal criterion.

\subsection{Problem Formulation}
Given a start state $q_{t_0} \in \mathcal{Q}^{valid}$, we aim to find a sequence of $K$ robot controls $\tau = \{u_{t_0}, \ldots, u_{t_k}, \ldots, u_{t_K}\}$ such that:
\begin{itemize}
    \item The end state arrives at a configuration inside the goal region: $\Gamma(q_{t_K}, u_{t_K}) \in \mathcal{Q}_G$; and
    \item All the intermediate states along the plan are valid, i.e., $\forall k: \; q_{t_k} \in \mathcal{Q}^{valid}$.
\end{itemize}

However, due to the modeling inaccuracies, physical and perception uncertainties, the control sequence $\tau$ will likely result in states different from what the planner predicted and cause task failures. As will be described below, in this work, we reformulate the problem by iteratively finding segments of controls $\overline{\tau} \subset \tau$, and interleave planning and robot execution between the control segments to close the manipulation loop, so that the system state is progressively transitioned towards the goal region $\mathcal{Q}_G$.

\section{Kinodynamic Manipulation Planning with Dynamic Horizons}
\label{sec:method}

To address the problem defined in Sec.~\ref{sec:problem}, we base our approach on the sampling-based kinodynamic RRT (kdRRT) framework \cite{lavalle2001randomized}, as it provides the ability to explore large high-dimensional state spaces, and can be easily integrated with any physics models or simulators to facilitate the generalization to various tasks. 

\begin{figure}[t]
    \centering
        \vspace{0.2cm}
        \includegraphics[width=0.49\columnwidth]{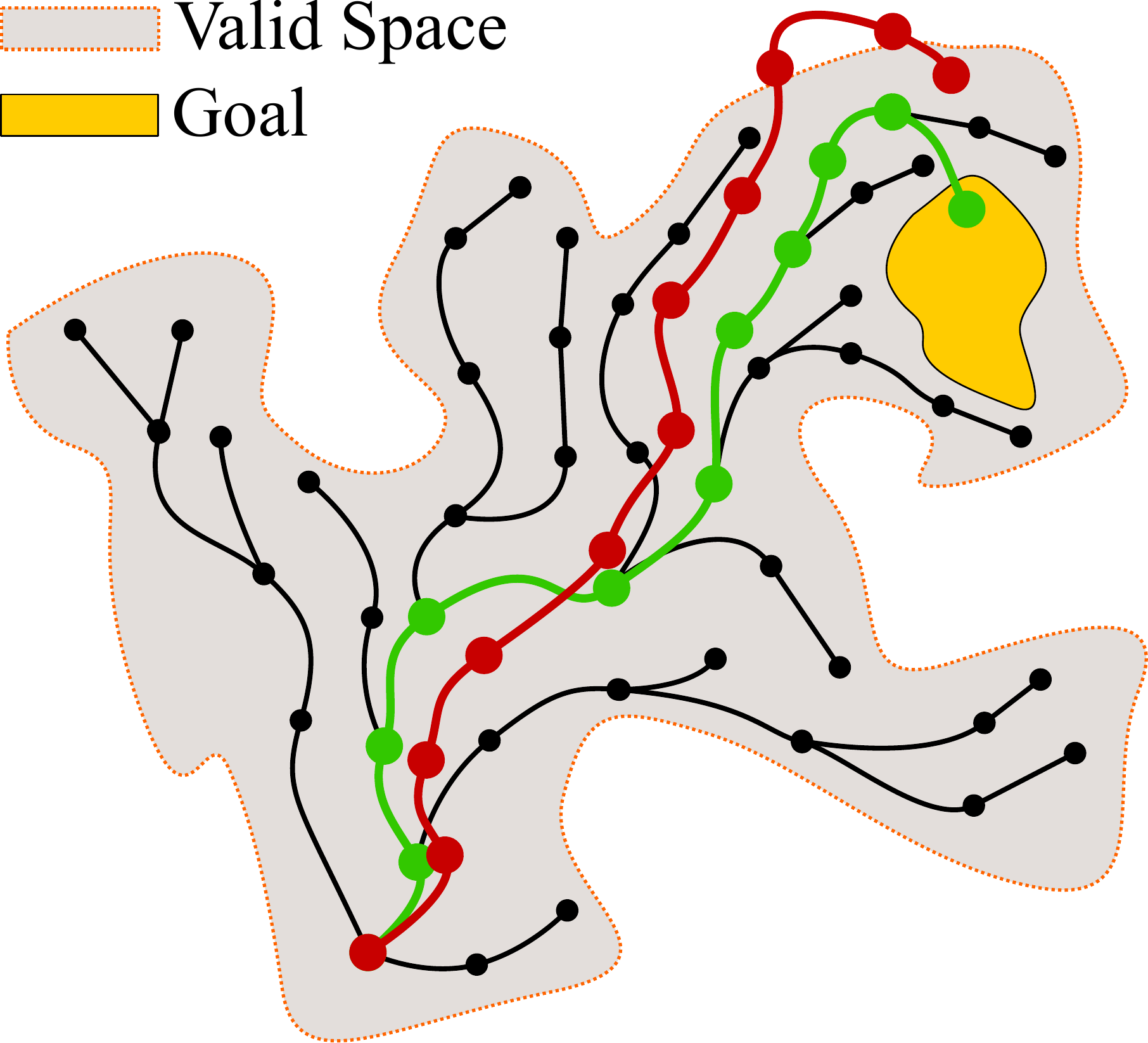}
        \includegraphics[width=0.49\columnwidth]{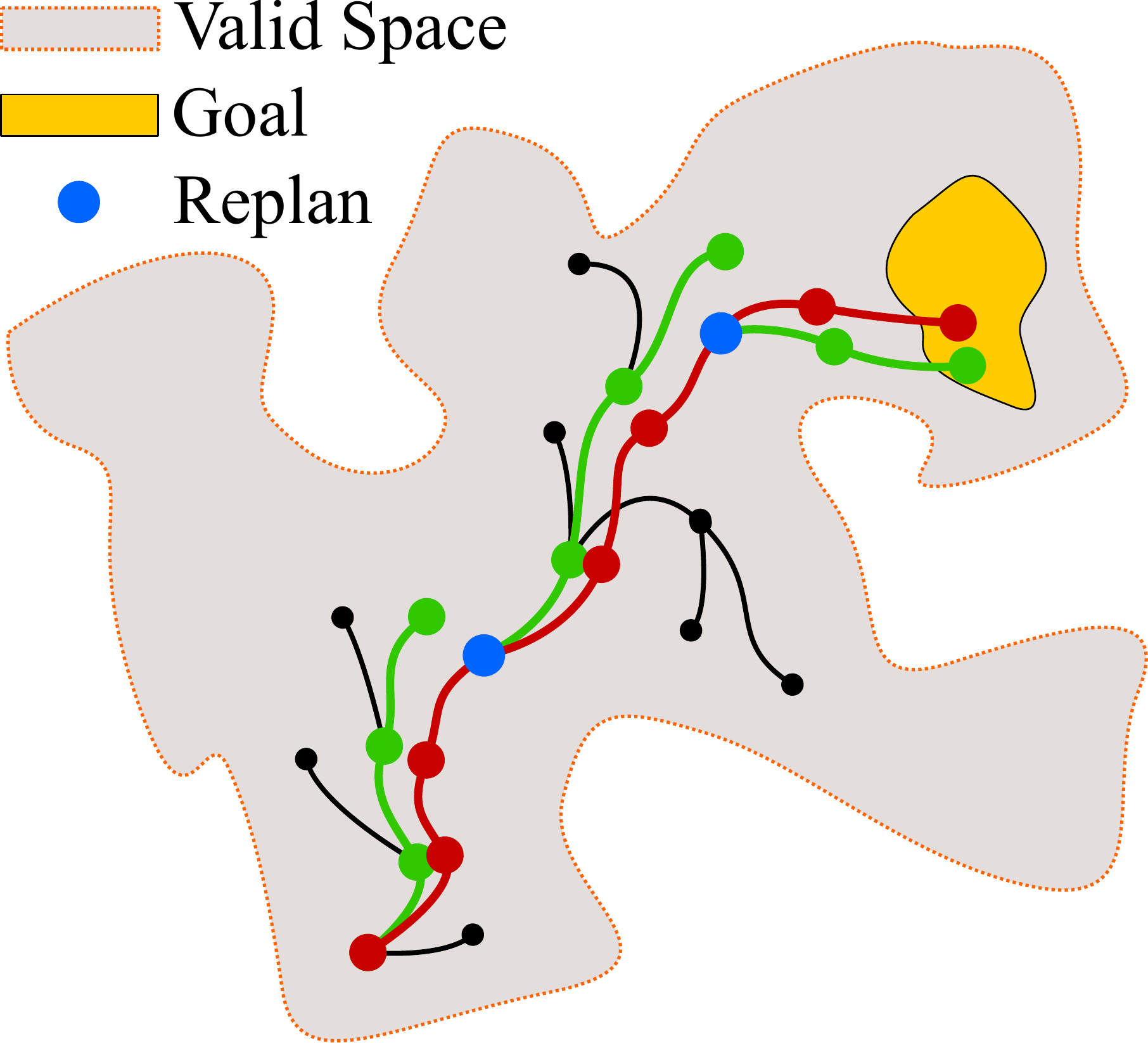}\\
        \vspace{0.1cm}
        \includegraphics[width=0.49\columnwidth]{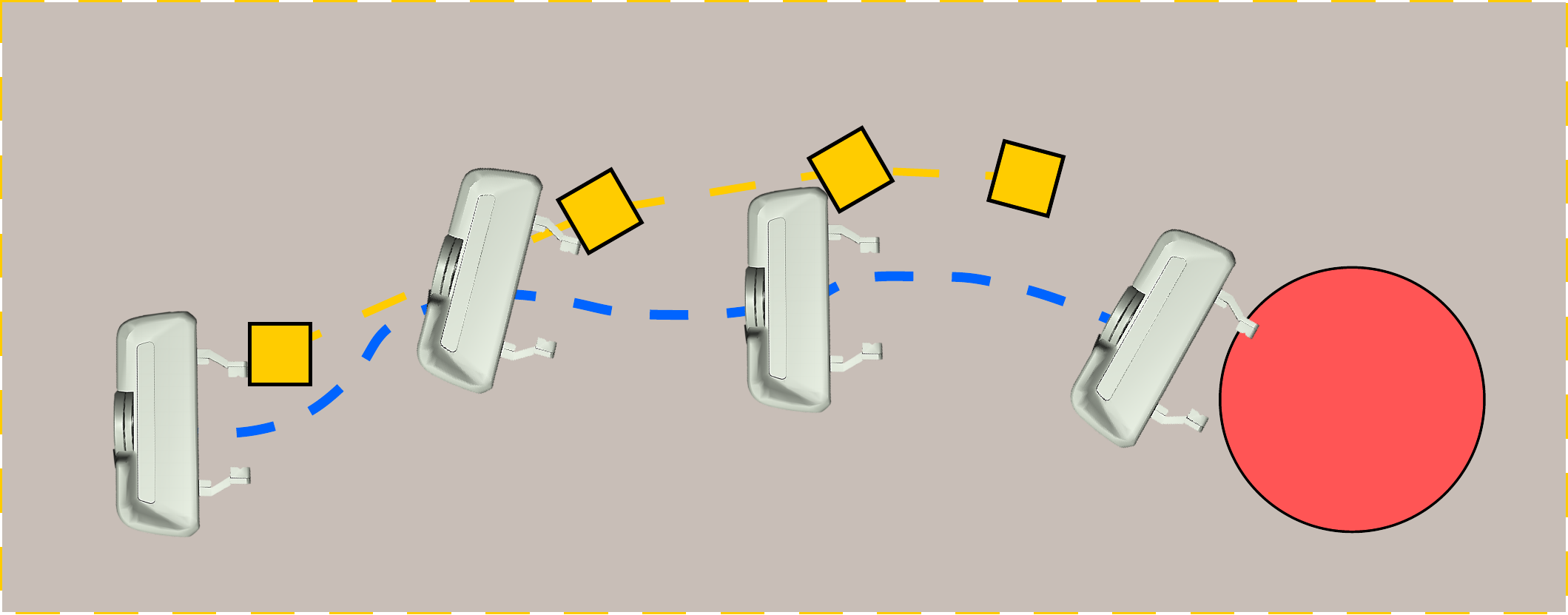}
        \includegraphics[width=0.49\columnwidth]{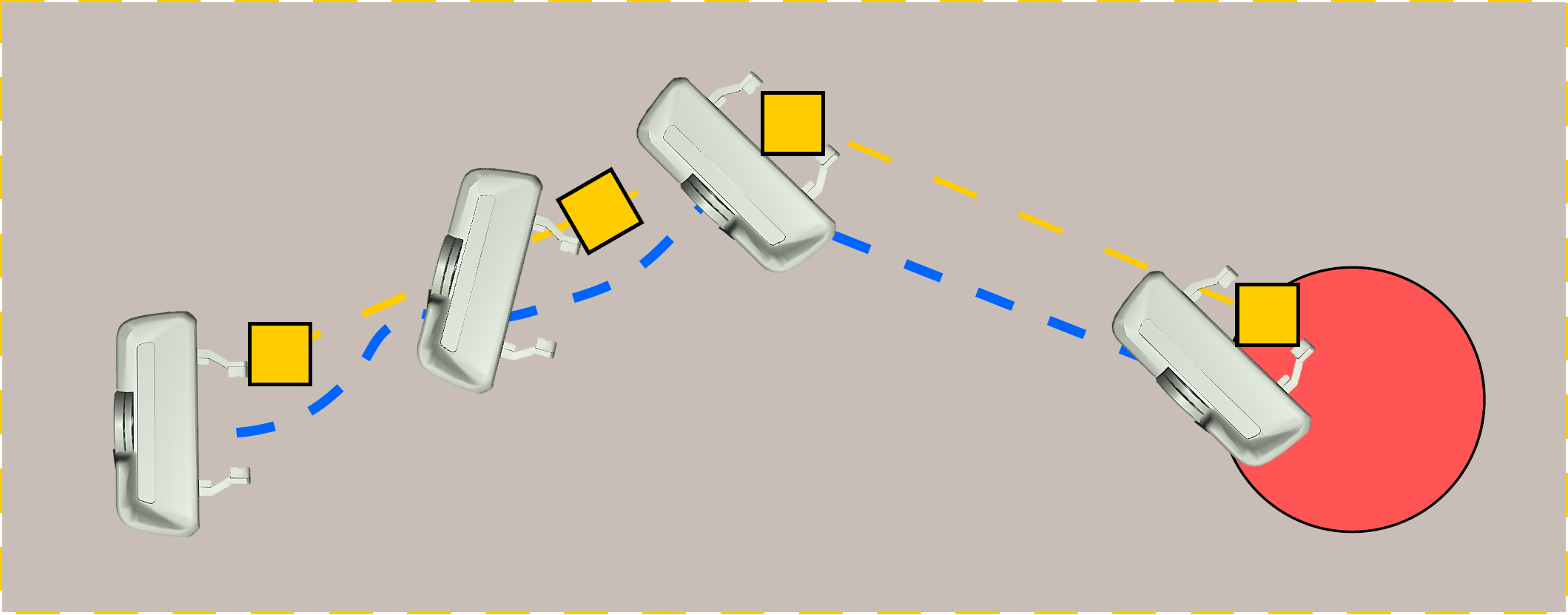}
    \caption{A schematic plot of dhRRT (Right), detailed in Sec.~\ref{sec:dhrrt}, compared to kdRRT (Left). \emph{Left}: The kdRRT extensively explores the entire valid space, and the execution (red) does not lead to the goal due to uncertainties, even if it finds a feasible plan (green). \emph{Right}: The dhRRT progressively grows its tree towards the goal, with replanning along the way, and finally reaches the goal.}
    \label{fig:schematic}
    \vspace{-0.5cm}
\end{figure}

Different from purely geometry-based motion planning algorithms, which explore the search space by sampling random state configurations and connect them to the search tree via linear interpolation, kinodynamic planning is much more complex due to the highly nonlinear system dynamics. 
%
%
In kinodynamic planning, given any sampled $q_{rand}$, one needs to find its nearest node, $q_{near}$, in the search tree, randomly sample a set of $M$ controls at $q_{near}$, $\{u^1, \ldots, u^M\}$, and then selects one $u^*$ to expand the search tree, such that the distance between $\Gamma(q_{near}, u^*)$ and $q_{rand}$ is minimized. This strategy of tree expansion can be computationally very expensive, especially for rearrangement problems involving multiple objects and many concurrent contacts, since adding each node into the tree requires $M$ times of physics calculation. Therefore, it could take up to several minutes to find a motion solution by expanding the tree very extensively in $\mathcal{Q}^{valid}$ \cite{king2015nonprehensile}. In addition, as illustrated in Fig.~\ref{fig:schematic}, given a motion solution (green), the real-world execution (red) can deviate due to the physical uncertainties. 

Another major challenge of such problems is the motion constraints. Although a robot arm can move in $SE(3)$, the motion of its end-effector has to be constrained within $\mathcal{W}$ to rearrange the objects, although sometimes teleport can be achieved via free-space motions. Therefore, control sampling needs to be constrained in a task manifold. For this, specialized planners can be used \cite{bereson09, kingston2019}. In this work, controls $v \in se(2)$ are sampled for the end-effector only, and are subsequently converted by Jacobian-based local projections, $\Call{JocobianProjection}{v}$, to check if any $u \in \mathcal{U}$ can realize the sampled controls, as detailed in Sec.~\ref{sec:jocobian}.

\subsection{Planning with Dynamic Horizons}
\label{sec:dhrrt}

To address the aforementioned challenges, we propose a kinodynamic RRT-based framework by incorporating Dynamic Horizon control in the planning process, and we term it as $dhRRT$.

For this, instead of explicitly defining goal configurations as in traditional settings, we require the task to be represented by a heuristic function: $h: \mathcal{Q}^{valid} \mapsto \mathbb{R}$, such that when the state of the system moves closer to the task goal, $h(\cdot)$ will decrease and will drive the system towards the goal region to eventually complete the task. In general, to guarantee that a search algorithm will be able to find an optimal solution for the task of $h(\cdot)$, the heuristic function has to be admissible and monotonic \cite{russell2002artificial}, meaning that $h(\cdot)$ should never overestimate the cost and should monotonically decrease as the state moves closer to the goal. However, it is infeasible to always prove or guarantee these two criteria due to the complexity of problems in the real world. In practice, since most rearrangement-based planning problems can be defined with simple cost-decreasing functions, non-optimal heuristics can also successfully drive the search to find solutions, although without optimality guarantees. As will be described in Sec.~\ref{sec:app} and shown with experiments, our planning framework can easily integrate such functions to achieve various tasks.

While we grow the search tree similarly to kdRRT algorithms, $h(\cdot)$ can be used to inform us about the planning progress. Given a search tree $T$ rooted at the start state, every node added in the tree represents a state $q_t$, with its inward edge representing a control $u_t$ that transitioned the state to $q_t$. During the tree expansion, we monitor the progress and will execute the current best control segment once good enough progress, determined by a threshold $p \in \mathbb{R}$, can be made by a leaf node. 

\begin{algorithm}
\caption{The dhRRT algorithm}
\small
    \begin{algorithmic}[1]
        \Require Start state $q_{t_0}$, goal region $\mathcal{Q}_G(\cdot)$, heuristic $h(\cdot)$, progress threshold $p$, tree limit $D_{max}$
        \Ensure Control sequence $\tau$
        \State $T \gets \{nodes=\{(q_{t_0}, 0)\}, edges=\emptyset\}$
        \State $\tau \gets \{\}$, $q^* \gets Null$
        \While{$\Call{Time.Available}{\null}$}
            \State $T \gets \Call{ExpandTree}{T}$
            \Comment{Alg.~\ref{alg:expandtree}}
            \State $\tau \gets \Call{EvaluateProgress}{T, h, p, D_{max}}$
            \Comment{Alg.~\ref{alg:evaluateprogress}}
            \If{$\tau \neq \{\}$}
                \State $ q^* \gets \Call{ExecuteControls}{\tau}$
                \Comment{Observe Real State}
                \If{$q^* \in \mathcal{Q}_G$}
                \Comment{Task Complete}
                \State \Return
                \EndIf
                \State $T \gets \{nodes=\{(q^*, 0)\}, edges=\emptyset\}$
                \State $\tau \gets \{\}$, 
                \State $q^* \gets Null$
                \State \textbf{continue}
            \EndIf
        \EndWhile
    \end{algorithmic}
    \label{alg:dhrrt}
\end{algorithm}

As such, our planning is controlled by a \emph{horizon} that dynamically changes in terms of the current state and motions. After each execution, our system observes the current state, which is likely to be different from the plan, and then repeats this procedure with different dynamically determined horizons, until the goal is reached. Note that, sometimes the robot can freely move around for a while without touching any object, hence making no positive progress. This is especially likely when the robot needs to relocate itself before making any rearrangement. To allow such motions without enforcing the robot to manipulate at every step, as well as facilitating random trap-escaping actions, the planning horizon is further limited by a threshold, $D_{max}$, of the maximum tree depth. If not enough progress can be made when $D_{max}$ is reached, our algorithm will execute the best solution so far. The algorithm is summarized in Alg.~\ref{alg:dhrrt}.

As illustrated in Fig.~\ref{fig:schematic}, rather than planning and executing the entire control sequence, our approach progressively transitions the system state towards the goal region with control segments, while observing the state in the real world after every $\Call{ExecuteControls}{\cdot}$, making it possible to close the manipulation loop to deal with errors along the path. In practice, the dynamic horizon threshold $p$ can be determined in terms of the expected magnitude of physical uncertainties, as well as the granularity of the physics models. By setting $p$ to a smaller value, the system will be more reactive, however, less efficient in finding solutions. Meanwhile, the tree depth limit $D_{max}$ can be set to smaller values to avoid getting trapped in local optimum via more random local motions, while a larger $D_{max}$ can allow more aggressive dynamic horizon control.

\begin{algorithm}[h]
\caption{ExpandTree($\cdot$)}
\small
    \begin{algorithmic}[1]
        \Require Current motion tree $T$
        \Ensure Expanded tree $T$
        \State $q_{rand} \gets \Call{SampleState()}{}$
        \State $q_{near} \gets \Call{FindNearest}{T, q_{rand}}$
        \For{$i = 1, ..., M$}
            \State $v_i \gets \Call{SampleControl()}{}$
            \Comment{In $se(2)$}
            \State $q_i \gets \Gamma(q_{near}, v_i)$
            \Comment{State Transition}
        \EndFor
        \State $(q_{new}, v^*) \gets \mathrm{\arg\min}_{(q_i, v_i)} \Call{Distance}{q_i, q_{rand}}$
        \State $u^* \gets \Call{JocobianProjection}{v^*}$
        \Comment{Sec.~\ref{sec:jocobian}}
        \If{$u^* \neq Null$}
            \State $T.\Call{AddNode}{q_{new}}$
            \Comment{Expansion}
            \State $T.\Call{AddEdge}{(q_{near}, q_{new}), u^*}$
        \EndIf
        \State \Return $T$
    \end{algorithmic}
    \label{alg:expandtree}
\end{algorithm}

\begin{algorithm}[h]
\caption{EvaluateProgress($\cdot$)}
\small
    \begin{algorithmic}[1]
        \Require Current motion tree $T$, heuristic $h(\cdot)$, progress threshold $p$, tree limit $D_{max}$
        \Ensure Control sequence $\tau$
        \State $q_{new} \gets T.\Call{GetLatestNode}{\null}$
        \State $\tau \gets \{\}$
        \If{$q_{new} \in \mathcal{Q}_G$}
        \Comment{Goal Reached}
            \State $\tau \gets \Call{ExtractControls}{T, q_{new}}$
        \ElsIf{$h(T.\Call{GetRoot}{\null}) - h(q_{new}) > p$}
        \Comment{Horizon}
            \State $\tau \gets \Call{ExtractControls}{T, q_{new}}$
        \ElsIf{$T.\Call{GetDepth()}{} = D_{\max}$}
        \Comment{Depth Limit}
            \State $q^\prime \gets \mathrm{\arg\min}_{q \in T.\Call{GetLeaves}{\null}} h(q)$
            \State $\tau \gets \Call{ExtractControls}{T, q^\prime}$
        \EndIf
        \State \Return $\tau$ 
    \end{algorithmic}
    \label{alg:evaluateprogress}
\end{algorithm}

\subsection{Jacobian-based Motion Projection}
\label{sec:jocobian}

Since we sample robot controls in the end-effector's velocity space in $se(2)$ to ensure the generated motions are constrained to the workspace, we need to project the controls to the robot's control space $\mathcal{U}$ to enable real robot executions. As indicated by the function $\Call{JacobianProjection}{\cdot}$ in Alg.~\ref{alg:expandtree}, for every new state $q_{new}$ to be added in the tree, we check whether the associated control $v^*$ can be projected to a valid $u^*$ to transition the state from $q_{near}$ to $q_{new}$.

Note that, as Jacobian matrix can constantly change while a control is being applied over a duration $[0, D]$, the Jacobian-based projection needs to be conducted continuously throughout the transition from $q_{near}$ to $q_{new}$, and a constant robot end-effector control $v^*$ can be generally projected to a smooth trajectory in $\mathcal{U}$. In practice, we address this by sufficiently discretizing the control duration with a small interval $\Delta t$, and then calculate the control $u^*_i$ for each intermediate state $q_i$.

Given a state $q_i$, its Jacobian matrix $J_i = \Call{Jacobian}{q_i^R}$ is calculated based on the current robot configuration $q_i^R \in \mathcal{Q}^R$. The control $u^*_i$ is then obtained by $u_i = J_i^\dag \cdot v^*$. While we iterative over $q_i$, we can determine that a control $u^*_i$ is invalid if: 1) the resulted robot configuration is invalid; or 2) the manipulability of the robot configuration, calculated by $\sqrt{\det J_i J_i^T}$ \cite{yoshikawa1985manipulability}, is smaller than a threshold, indicating that the robot is going to hit its singularity. If every intermediate projection is valid, $\Call{JacobianProjection}{\cdot}$ will return by composing a control trajectory $u^*$ based on all the intermediate $u^*_i$, and will otherwise return $Null$.

\section{Example Applications}
\label{sec:app}

To evaluate our framework, we task the robot with $3$ different rearrangement-based manipulation tasks in clutter, as exemplified in Fig.~\ref{fig:first}.

\subsubsection{Grasping}

For grasping a target object in clutter, the robot needs to rearrange the surrounding objects so that the gripper can reach a stable pre-grasp pose. The major challenge of this task is that, while the surrounding objects are being rearranged, the target object is simultaneously  moved by object-object interactions. The task goal is achieved when the center of the target object $(x^o, y^o)$ is inside the area between the two fingers, denoted as $\mathcal{G}^R$, and the orientation of the gripper is roughly aligned with a feasible grasping angle. Formally, the goal criterion is:
\begin{equation}
    (x^o, y^o) \in \mathcal{G}^R \land \min_{\alpha \in \mathcal{A}} \lvert \theta^R - \alpha \rvert \leq \epsilon_\alpha
\end{equation}
where $\theta^R$ is the orientation of the gripper, $\mathcal{A}$ is the set of feasible grasping angles, and $\epsilon_\alpha > 0$ is a threshold in radians for which we set to be $0.2$ in all experiments. 

The heuristic function used by our dhRRT planner for grasping encourages the gripper to approach the target object. Let us denote the state of the gripper as $(x^R, y^R, \theta^R) \in SE(2)$, we define the grasping heuristic function $h_g$ to take the following simple form:
\begin{equation}
\begin{aligned}
    h_g(q) = & w_d \cdot \sqrt{(x^R - x^o)^2 + (y^R - y^o)^2}\\
    + & w_\alpha \cdot \lvert \theta^R - \mathrm{atan}(y^o - y^R, x^o - x^R) \rvert
\end{aligned}
\end{equation}
where $w_d$, $w_\alpha$ are weighting factors and set to be $w_d = 0.7$, $w_\alpha = 0.3$ in all experiments.

\subsubsection{Relocating}

The relocating task for the robot is to push the target object to a circular goal region $\mathcal{G}$ centered at $(x_\mathcal{G}, y_\mathcal{G})$ with a radius of $0.1$m. The goal criterion is:
\begin{equation}
    (x^o, y^o) \in \mathcal{G}
\end{equation}
This is a difficult task since the target object is not necessarily reachable by the gripper, and its motion is indirectly determined by all other objects. The heuristic function used by our dhRRT planner is simply defined by the summation of the distance between the target object and the gripper, and the distance between the target object and the goal region:
\begin{equation}
\begin{aligned}
    h_r(q) = &\sqrt{(x^o - x^R)^2 + (y^o - y^R)^2}\\
    + &\sqrt{(x^o - x_\mathcal{G})^2 + (y^o - y_\mathcal{G})^2}
\end{aligned}
\end{equation}

\subsubsection{Sorting}

The sorting task is to rearrange all the movable objects to separate them into different classes, which are represented by different colors in our experiments. We denote by $L$  the number of object classes, and by $\mathbf{CH}_i (q) \subseteq \mathbb{R}^2$, $i \in \{1, ..., L\}$, the convex hull containing all the objects of the $i$-th class in state $q \in \mathcal{Q}^{valid}$, then the goal criterion of sorting is satisfied if all classes have at least a distance $\epsilon_d > 0$ from each other. Formally, $\forall i,j \in \{1,\ldots, L\}$:
\begin{equation}
\begin{aligned}
   \min_{i \neq j} \Call{Distance}{\mathbf{CH}_i (q), \mathbf{CH}_j (q)} > \epsilon_d
\end{aligned}
\end{equation}

Our dhRRT planner uses a heuristic function similar to the reward signals in our previous work in \cite{song2020multi} for sorting. Intuitively, a state where the objects are placed closer to each other for the same class and further apart for different classes will receive a lower cost value.

%

\section{Experiments}
\label{sec:experiments}

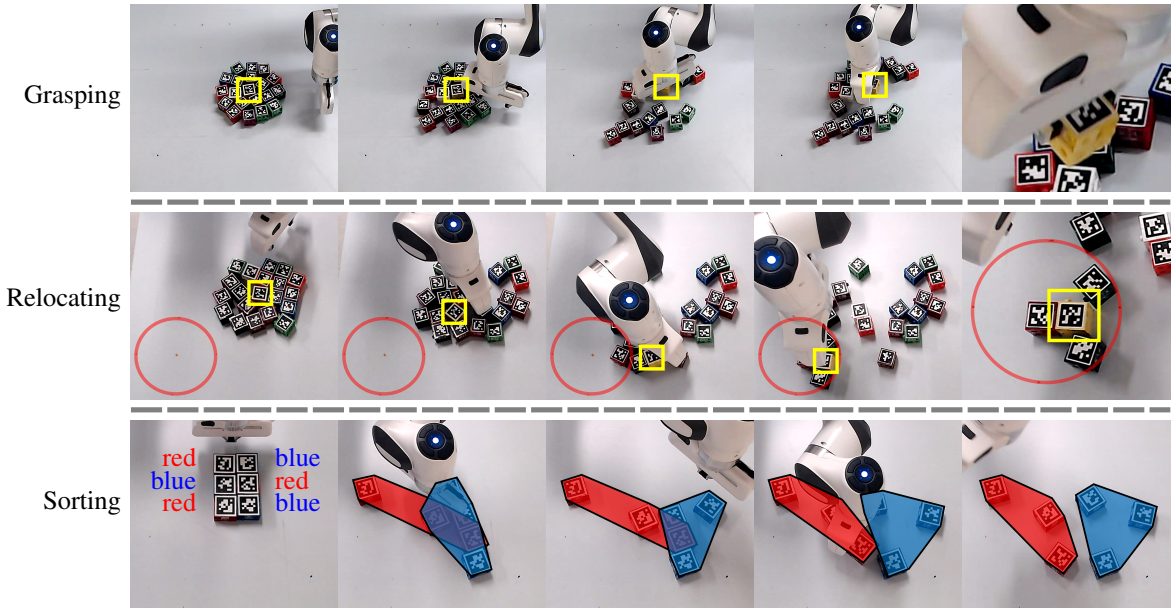
\begin{figure*}[t]
\centering
    \begin{tikzpicture}
        \node[anchor=south west,inner sep=0] at (0,0){\includegraphics[width=1.6\columnwidth]{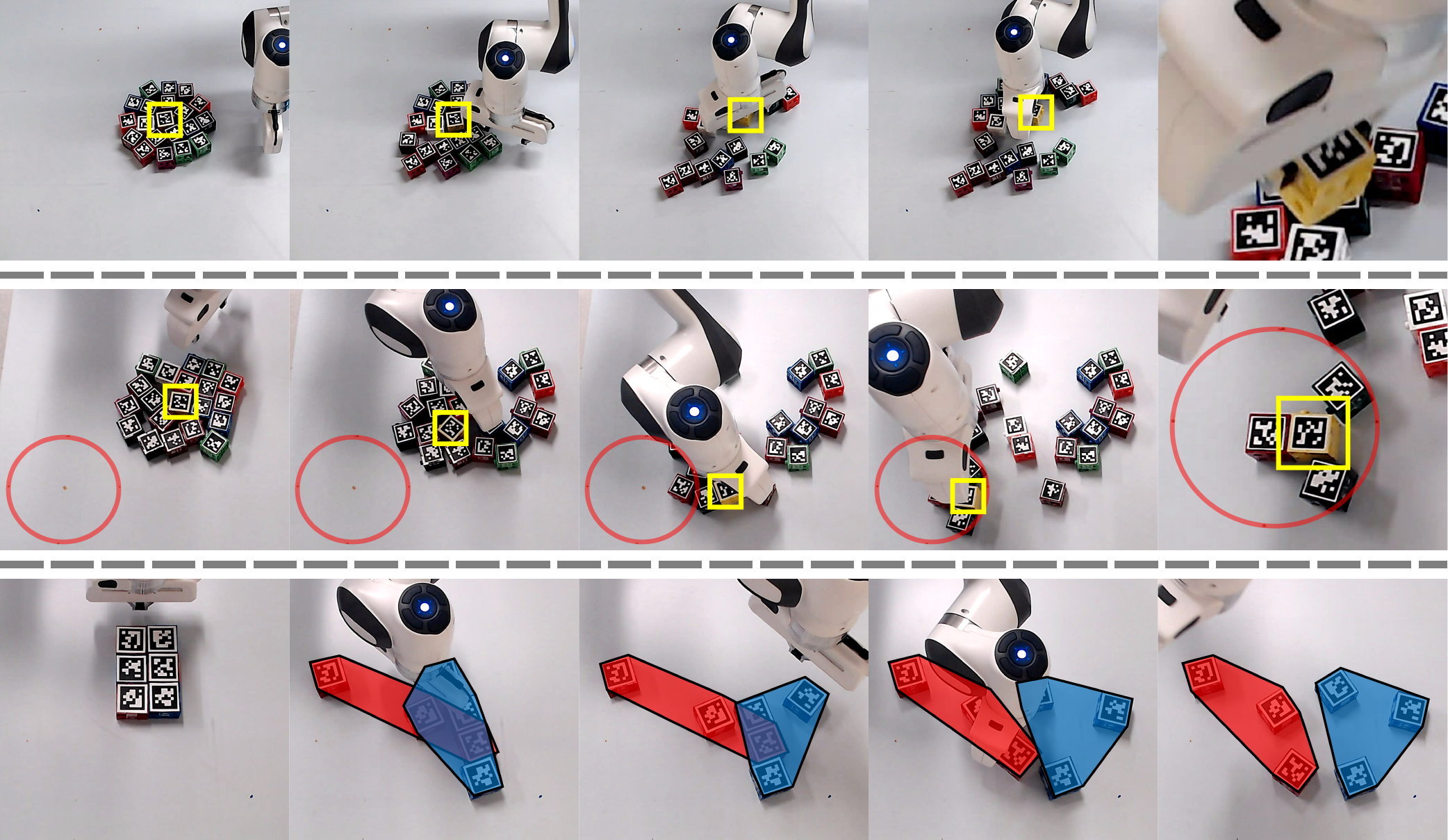}};
        
        \node[anchor=east, align=right] at (0, 6.8) {Grasping};
        \node[anchor=east, align=right] at (0, 4.1) {Relocating};
        \node[anchor=east, align=right] at (0, 1.4) {Sorting};
        
        \node[anchor=east, align=right, text=red] at (1, 1.4) {red};
        \node[anchor=east, align=right, text=blue] at (1, 1.7) {blue};
        \node[anchor=east, align=right, text=red] at (1, 2) {red};
        
        \node[anchor=west, align=left, text=blue] at (1.8, 1.4) {blue};
        \node[anchor=west, align=left, text=red] at (1.8, 1.7) {red};
        \node[anchor=west, align=left, text=blue] at (1.8, 2) {blue};
    \end{tikzpicture}
\caption{Real-world experiments on $3$ rearrangement-based manipulation tasks: grasping, relocating, and sorting. The target object is highlighted by yellow boxes for grasping and relocating, and the convex hulls for object classes are shown in blue and red for sorting. }
\label{fig:real_world_tasks}
\end{figure*}

With the three tasks defined in Sec.~\ref{sec:app}, we evaluate the proposed framework from three aspects relevant to real-world challenges. First, by increasing the size of the object clutter (number of objects), we test and report the success rate and planning efficiency of the planner. Second, we evaluate the robustness against inaccurate models with quantified model granularities. Finally, we challenge the planner by introducing nondeterministic physics to evaluate its reactivity. Our experiments were conducted both with a real Franka Emika Panda robot and in the MuJoCo simulator \cite{todorov2012mujoco}. All objects were tracked via AprilTags \cite{olson11}.

In addition, we implemented two baseline algorithms to compare with the proposed dhRRT approach. First, we implemented a kdRRT algorithm modified from \cite{king2015nonprehensile}, and for a fair comparison, we replaced the physics model with the MuJoCo simulator, and we do not limit the number of concurrent contacts. Second, we enable kdRRT with replanning, termed as r-kdRRT, by observing the end states after every execution, and will trigger replanning if the goal is not reached. In all experiments, the controls were sampled in the robot gripper's velocity space. The linear velocity was bounded by $[-0.2, 0.2]m/s$ in simulation, but by $[-0.1, 0.1]m/s$ in the real world for better safety. The angular velocity was all bounded by $[-1, 1]rad/s$. The control duration was fixed to $0.2$ seconds (grasping, relocation) and $0.4$ seconds (sorting). In addition, all reported planning times were calculated from the successful runs only, and all the time budgets were set for planning only, excluding the execution time.

\subsection{Efficiency and Robustness}

For this part of our evaluation, we conducted only real-world experiments as the planning efficiency will be similar to simulation-based experiments, but the system's robustness can be more realistically challenged in the real world. Example executions for the $3$ tasks are shown in Fig.~\ref{fig:real_world_tasks}. For the grasping and relocating tasks, we used $N=10$ and $N=20$ objects, with one of them being the target object in each task. The sorting task used $6$ objects and $2$ classes: $3$ blue objects and $3$ red objects. For all experiments with our algorithm and the baseline algorithms, the time budget was set to $60$ seconds, accumulated through the process if replanning was needed.

\begin{figure}[t]
\setlength{\tabcolsep}{5pt}
\centering
\footnotesize
\begin{subtable}[h]{\linewidth}
    \centering
    \caption{Grasping}
    \vspace{-5pt}
    \begin{tabular}{c|c|| c c c}
        \hline
        Scene & Metric & kdRRT & r-kdRRT & dhRRT\\
        \hline
        \multirow{2}{*}{N = 10} & Success Rate & \: 1 / 10 & \: 6 / 10 & \textbf{10 / 10}\\
        & Time (seconds) & 13.0 $\pm$ 0.0 & 22.5 $\pm$ 16.7 & \textbf{ 9.8 $\pm$ 2.5}\\
        \hline
        \multirow{2}{*}{N = 20} & Success Rate & \: 0 / 10 & \: 2 / 10 & \textbf{\: 8 / 10}\\
        & Time (seconds) & - & 20.1 $\pm$ 8.6 & \textbf{11.0 $\pm$ 4.3}\\
        \hline
    \end{tabular}
\end{subtable}
\vspace{1pt}

\begin{subtable}[h]{\linewidth}
    \centering
    \caption{Relocating}
    \vspace{-5pt}
    \begin{tabular}{c|c|| c c c}
        \hline
        Scene & Metric & kdRRT & r-kdRRT & dhRRT\\
        \hline
        \multirow{2}{*}{N = 10} & Success Rate & \: 3 / 10 & \: 4 / 10 & \textbf{\: 9 / 10}\\
        & Time (seconds) & 33.6 $\pm$ 16.6 & 40.1 $\pm$ 18.2 & \textbf{15.2 $\pm$ 4.9}\\
        \hline
        \multirow{2}{*}{N = 20} & Success Rate & \: 1 / 10 & \: 2 / 10 & \textbf{\: 9 / 10}\\
        & Time (seconds) & 34.9 $\pm$ 0.0 & 35.4 $\pm$ 0.5 & \textbf{18.1 $\pm$ 11.1}\\
        \hline
    \end{tabular}
\end{subtable}
\vspace{1pt}

\begin{subtable}[h]{\linewidth}
    \centering
    \caption{Sorting}
    \vspace{-5pt}
    \begin{tabular}{c|| c c c}
        \hline
        Metric & kdRRT & r-kdRRT & dhRRT\\
        \hline
        Success Rate & \: 0 / 10 & \: 0 / 10 & \textbf{4 / 10}\\
        Time (seconds) & -- & -- & \textbf{28.3 $\pm$ 16.4}\\
        \hline
    \end{tabular}
\end{subtable}
\caption{Experiment results of the $3$ real-world tasks.}
\label{tab:real_eval}
\vspace{-0.6cm}
\end{figure}

The experiment results are reported in Fig.~\ref{tab:real_eval}. We can see that while both kdRRT and r-kdRRT could barely achieve $50\%$ of success rates in grasping and relocating, our dhRRT has succeeded in more than $90\%$ of tests, with only failure case happening due to one of the objects being pushed outside of the workspace due to physical uncertainties. It is evident that the dynamic horizon significantly improved the robustness by closing the manipulation loop, and enabled progressive manipulation while observing the real-world transitions. Moreover, we observe that dhRRT is much more efficient than the baseline algorithms. This is because: 1) dhRRT can more adaptively focus on task-relevant subspaces of the problem; and 2) it avoids making large trajectory deviations and requires less replanning effort compared with r-kdRRT.

For the sorting task, however, while none of the baseline algorithms could solve it at all, dhRRT achieved $40\%$ success rate. This is due to the much higher complexity of the sorting problem, which does not have a single target object, and the goal is achieved only when \emph{all objects} are relatively reconfigured in certain ways. Under the time budget, and without training or carefully designing problem-specific heuristics, such as done in \cite{song2020multi, huang19}, as well as not being able to teleport the gripper in between of actions, our dhRRT was not able to provide good sorting performance in the real physical world, although we achieved a success rate of $90\%$ in simulation.

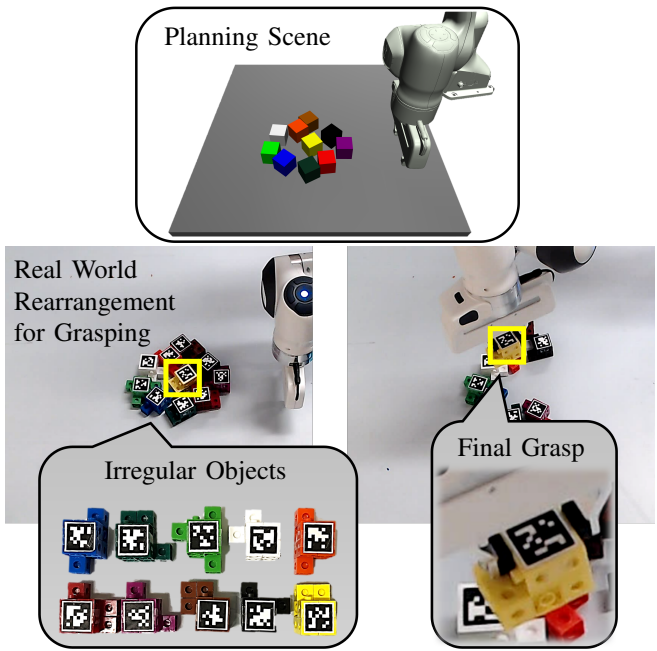
\begin{figure}[h]
    \centering
    \begin{tikzpicture}
        \node[anchor=south west,inner sep=0] at (0,0){\includegraphics[width=\linewidth]{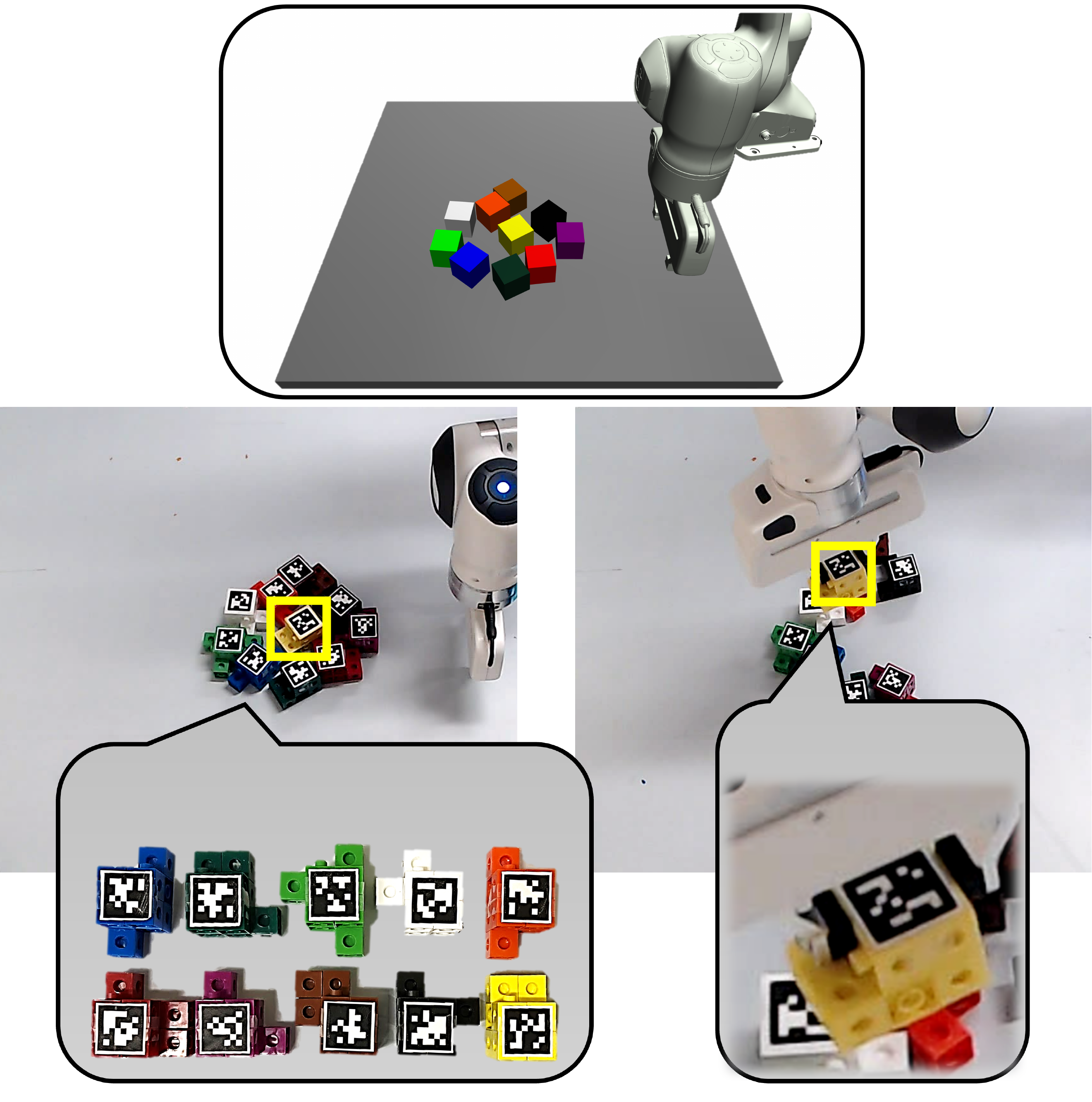}};
        
        \node[anchor=north west, align=left] at (2, 8.55) {Planning Scene};
        \node[anchor=west, align=left] at (0, 4.75) {Real World\\ Rearrangement \\for Grasping};
        \node[anchor=west, align=left] at (1.2, 2.5) {Irregular Objects};
        \node[anchor=west, align=left] at (5.9, 2.8) {Final Grasp};
    \end{tikzpicture}
    \caption{The robot is tasked to grasp the yellow object in a clutter. Despite the fact that our planner incorrectly models perfect cube shapes for all objects, it can successfully rearrange the irregular object shapes in the real world and complete the task via dynamic planning horizons.}
\label{fig:real_shape}
\vspace{-0.2cm}
\end{figure}

\begin{figure}[!t]
\footnotesize
\centering
    \begin{tabular}{c|| c c c}
        \hline
        Metric & kdRRT & r-kdRRT & dhRRT\\
        \hline
        Success Rate & \: 0 / 10 & \: 3 / 10 & \textbf{\:9 / 10}\\
        Time (seconds) & -- & 42.3 $\pm$ 9.9 & \textbf{8.6 $\pm$ 3.4}\\
        \hline
    \end{tabular}
\caption{Experiment results of real-world grasping tasks with inaccurate object models.}
\label{tab:real_shape}
\vspace{-0.2cm}
\end{figure}

\begin{figure}[!t]
    \centering
    \includegraphics[width=\linewidth]{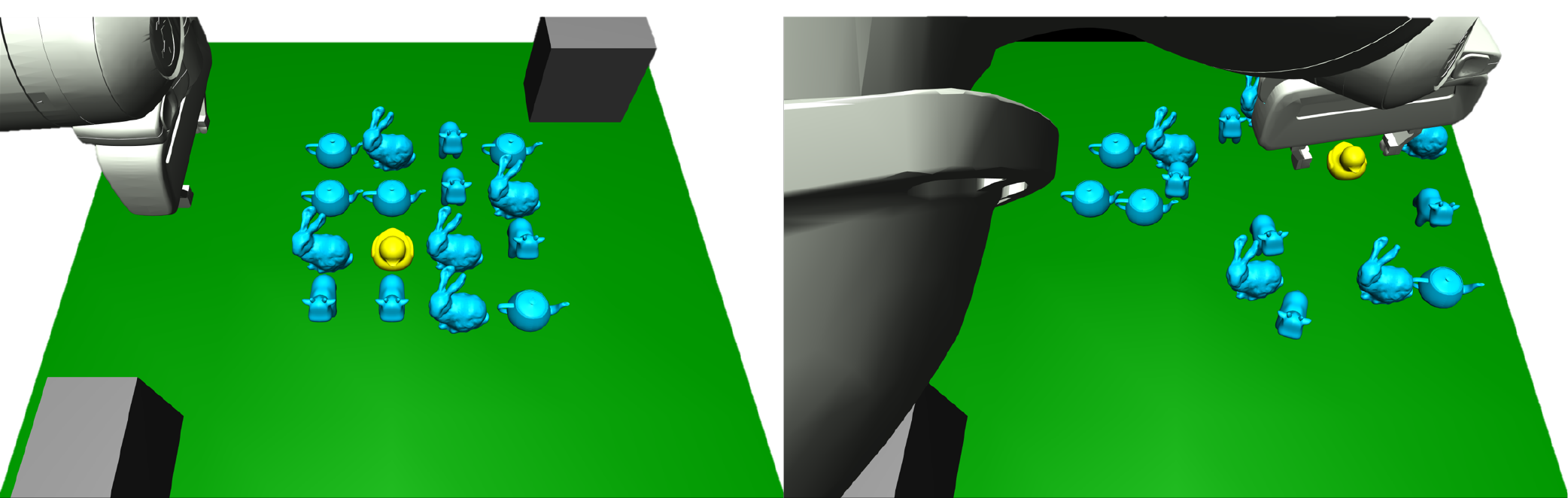}
    \caption{Experiment setup for grasping with resolution-reduced object models. The target object is shown in yellow and static obstacles are shown in gray. \emph{Left}: start configuration. \emph{Right}: a goal state reached by our dhRRT.}
\label{fig:mesh_grasp}
\vspace{-0.5cm}
\end{figure}

\subsection{Inaccurate Object Models}

In real-world applications, due to the perception limitations, we do not always have access to perfect object models. For example, the object models can be incorrectly estimated, or the resolutions of the models are not good enough to reflect real-world contact physics. As such, we designed two test cases to evaluate the algorithms' robustness against perception uncertainties in grasping tasks by: 1) using incorrect object models in planning; and 2) iteratively reducing the resolution of object models in planning. For case 1), we conducted real-world experiments using $10$ objects as exemplified in Fig.~\ref{fig:real_shape}. We tested case 2) in simulation to be able to access perfect and resolution-reduced object models. The time budget was set to $60$ seconds and $120$ seconds for case 1) and case 2) respectively.

The experiment results of case 1) are summarized in Fig.~\ref{tab:real_shape}. Due to the discrepancies between object models in planning and in the real world, kdRRT was never able to complete the task. Under this difficult setting, our dhRRT was able to succeed $9$ out of $10$ times, and the planning efficiency was almost not affected in comparison to the cases where accurate models were available. In addition, even if r-kdRRT was able to complete $3$ out of $10$ trials, the planning efficiency was, again, evidently lower than our dhRRT, indicating that the proposed dynamic horizons provided essential reactivity to the system for both improved efficiency and robustness.

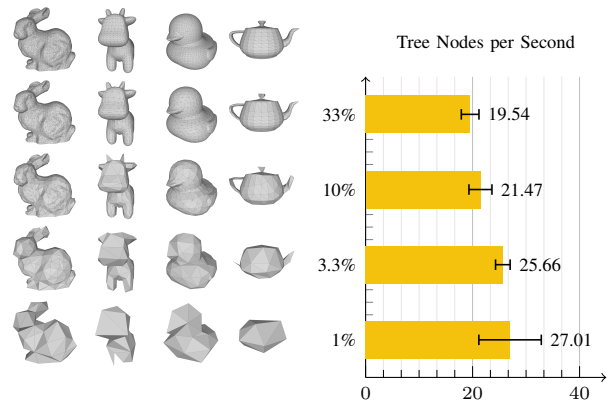
\begin{figure}[t]
    \centering
    \begin{minipage}[b]{0.45\linewidth}
        \includegraphics[width=\linewidth]{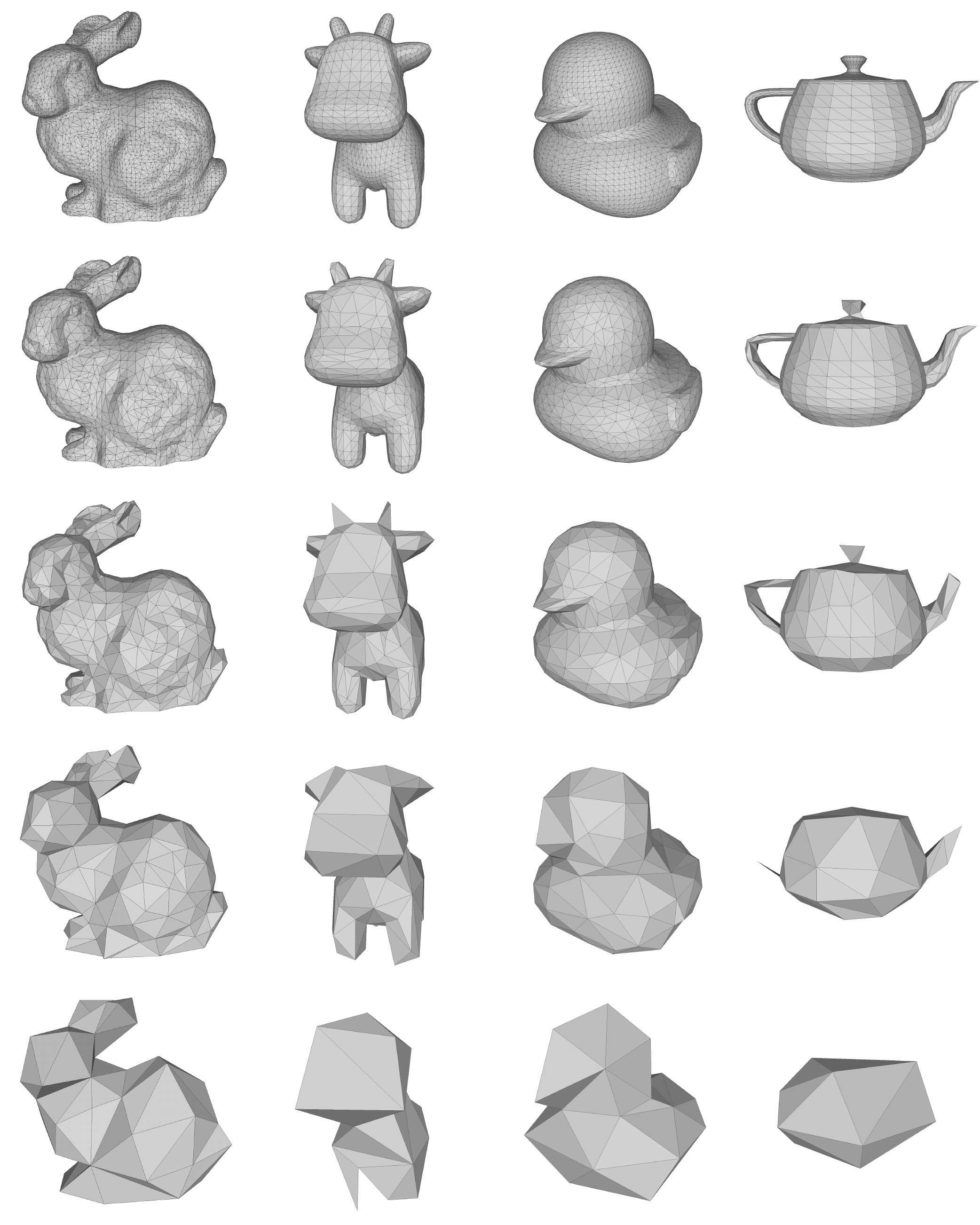}
        \vspace{0.5cm}
    \end{minipage}
    \begin{minipage}[b]{0.45\linewidth}
\begin{tikzpicture}

\definecolor{mycolor1}{rgb}{0.95686,0.76078,0.05098}%

\begin{axis}[
width=3.2cm,
height=4cm,
scale only axis,
axis x line*=bottom,
axis y line*=left,
legend columns=4,
grid=both,
grid style={line width=.1pt, draw=gray!20},
major grid style={line width=.2pt,draw=gray!50},
minor tick num=5,
ymajorgrids=false,
yminorgrids=false,
xmin=0, xmax=45,
xtick style={color=black},
ytick={4, 3, 2, 1},
yticklabels={33\%,10\%,3.3\%,1\%},
ymin=0.5, ymax=4.5,
axis background/.style={fill=white},
title=Tree Nodes per Second
]
\draw[draw=none,fill=mycolor1] (axis cs:0,3.75) rectangle (axis cs:19.5441530696655,4.25);
\draw[draw=none,fill=mycolor1] (axis cs:0,2.75) rectangle (axis cs:21.4697612312229,3.25);
\draw[draw=none,fill=mycolor1] (axis cs:0,1.75) rectangle (axis cs:25.6556666852237,2.25);
\draw[draw=none,fill=mycolor1] (axis cs:0,0.75) rectangle (axis cs:27.0106247317318,1.25);
\path [draw=black, semithick]
(axis cs:17.8939320900367,4)
--(axis cs:21.1943740492943,4);

\path [draw=black, semithick]
(axis cs:19.3189812823286,3)
--(axis cs:23.6205411801172,3);

\path [draw=black, semithick]
(axis cs:24.2921583848449,2)
--(axis cs:27.0191749856025,2);

\path [draw=black, semithick]
(axis cs:21.1739620697997,1)
--(axis cs:32.847287393664,1);

\addplot [semithick, black, mark=|, mark size=2, mark options={solid}, only marks]
table {%
17.8939320900367 4
19.3189812823286 3
24.2921583848449 2
21.1739620697997 1
};
\addplot [semithick, black, mark=|, mark size=2, mark options={solid}, only marks]
table {%
21.1943740492943 4
23.6205411801172 3
27.0191749856025 2
32.847287393664 1
};
\draw (axis cs:21.1943740492943,4) ++(0pt,0pt) node[
  anchor=west,
  text=black,
  rotate=0.0
]{\scriptsize 19.54};
\draw (axis cs:23.6205411801172,3) ++(0pt,0pt) node[
  anchor=west,
  text=black,
  rotate=0.0
]{\scriptsize 21.47};
\draw (axis cs:27.0191749856025,2) ++(0pt,0pt) node[
  anchor=west,
  text=black,
  rotate=0.0
]{\scriptsize 25.66};
\draw (axis cs:32.847287393664,1) ++(0pt,0pt) node[
  anchor=west,
  text=black,
  rotate=0.0
]{\scriptsize 27.01};
\end{axis}

\end{tikzpicture}
    \end{minipage}
    \vspace{-10pt}
    \caption{Object models used in the experiment shown in Fig.~\ref{fig:mesh_grasp}. \emph{Left}: The resolution-reduction level, determined by face reduction percentage, were $100\%$ (original), $33\%, 10\%, 3.3\%$, and $1\%$ from top to bottom. \emph{Right:} Number of nodes added in the tree per second in terms of the reduction rates.}
\label{fig:shape}
\end{figure}

The experiment setup for case 2) is illustrated in Fig.~\ref{fig:mesh_grasp}. We selected $4$ complex object shapes, shown in Fig.~\ref{fig:shape}, and iteratively reduced their resolutions to $33\%, 10\%, 3.3\%$, and $1\%$ to be used by planners. Note that, once a plan was generated for execution, the simulated execution used only the original object models. We can observe that, as reported in Fig.~\ref{tab:shape}, although the model reduction significantly affected the baseline algorithms, dhRRT still could complete the task with high success rates. Also, the planning time for dhRRT was much lower than the baseline approaches. Notice that, the planning time for the baseline algorithms was occasionally faster than dhRRT due to their small number of successes and randomness in statistics. Furthermore, Fig.~\ref{fig:shape} reports the tree expansion efficiency in terms of the number of added nodes per second. It is interesting to note that, although imperfect models enlarge the gaps between planning and the real world, simpler models can facilitate the planning efficiency. Therefore, in terms of the task requirements, in practice we can balance between the needed robustness and the modeling resolution of models to achieve higher efficiency.

\begin{figure}[!t]
\setlength{\tabcolsep}{3pt}
\footnotesize
\centering
    \begin{tabular}{c||c c c ||c c c}
        \hline
        Red.& \multicolumn{3}{c||}{Success Rate} & \multicolumn{3}{c}{Time (seconds)}\\
        \cline{2-7}
        Rate.& kdRRT & r-kdRRT & dhRRT & kdRRT & r-kdRRT & dhRRT\\
        \hline
        33 \% & \: 5/20 & 16/20 & \textbf{20/20} & \textbf{11.9$\pm$\:4.8} & 48.8$\pm$32.0 & 13.3$\pm$13.7\\
            
        10 \% & \: 3/20 & 12/20 & \textbf{17/20} & 28.7$\pm$12.5 & 34.2$\pm$28.5 & \textbf{17.6$\pm$10.4}\\
            
        3.3 \% & \: 2/20 & 11/20 & \textbf{15/20} & \: \textbf{2.2$\pm$  0.7} & 38.9$\pm$30.2 & 16.1$\pm$ 9.4\\
            
        \: \:1 \% & \: 1/20 & 12/20 & \textbf{17/20} & 24.4$\pm$  0.0 & 32.1$\pm$19.0 & \textbf{22.6$\pm$18.1}\\
        \hline
    \end{tabular}
\caption{Results of grasping tasks with resolution-reduced models.}
\label{tab:shape}
\end{figure}

\begin{figure}[!t]
\setlength{\tabcolsep}{3pt}
\centering
\footnotesize
    \begin{tabular}{c||c c c ||c c c}
        \hline
         & \multicolumn{3}{c||}{Success Rate} & \multicolumn{3}{c}{Time (seconds)}\\
        \cline{2-7}
        $\Delta t(s)$ & kdRRT & r-kdRRT & dhRRT & kdRRT & r-kdRRT & dhRRT \\
        \hline
        20 & 17/20 & 17/20 & \textbf{20/20} & 27.2$\pm$19.7 & 27.2$\pm$19.7 & \textbf{20.1$\pm$17.7}\\
            
        10 & 17/20 & 19/20 & \textbf{20/20} & 22.9$\pm$15.3 & 25.6$\pm$16.8 & \textbf{15.1$\pm$11.0}\\
            
        \: 5 & \: 7/20 & 16/20 & \textbf{19/20} & 26.5$\pm$18.9 & 40.7$\pm$23.5 & \textbf{19.5$\pm$22.8}\\
            
        \: 2 & \: 3/20 & 12/20 & \textbf{17/20} & 24.9$\pm$11.3 & 45.6$\pm$33.5 & \textbf{14.9$\pm$11.1}\\
        
        \: 1 & \: 3/20 & \: 6/20 & \textbf{15/20} & 13.2$\pm$ 7.0 & 15.8$\pm$ 7.4 & \textbf{11.6$\pm$ 8.0}\\
        
        0.5 & \: 0/20 & \: 9/20 & \textbf{17/20} & -- & 35.9$\pm$23.7 & \textbf{13.3$\pm$ 7.4}\\
        \hline
    \end{tabular}
    \caption{Results of grasping tasks under nondeterministic physics.}
\label{tab:interference}
\vspace{-0.2cm}
\end{figure}

\subsection{Nondeterministic Physics}

Another factor that challenges manipulation planning in the real world is the nondeterministic physics. For example, the object's friction coefficient against the table surface is not a constant resulting in different dynamics over the execution, or that the objects can be moved by external perturbations. Therefore, we designed an experiment to introduce random local perturbations to the objects during the execution of manipulation plans. In this experiment, we used $16$ cubes similar to the experiments shown in Fig.~\ref{fig:first}, and in every time interval $\Delta t$, we randomly select one object and assign it with a linear velocity of $0.4m/s$, and the goal is for the robot to grasp the target object. The time budget was set to $120$ seconds.

The results are reported in Fig.~\ref{tab:interference}. In this experiment, a shorter interval for applying random perturbations simulates a higher level of nondeterministic physics. Being consistent with other experiments, dhRRT outperformed both baseline planners in efficiency and robustness, which again shows that the proposed dhRRT planner can focus the planning procedure to more task-relevant subspaces, and facilitates a close-loop manipulation solution against physical uncertainties in executing motions plans.

\section{Conclusion}
\label{sec:conclusion}

We presented a kinodynamic manipulation planning framework for rearrangement-based manipulation problems. Based on efficient sampling-based planning, we proposed to monitor and dynamically adapt the kinodynamic planning horizons, and progressively transition the system states towards the goal region by interleaving planning and execution, which greatly enhanced the system's robustness. Using simple heuristics, we showed that our approach is not only able to focus the planning on task-relevant subspaces to significantly improve the planning efficiency, but also enables implicit definitions of manipulation goals, in contrast to many traditional goals defined by explicit configurations. 

With extensive experiments both in the real world and in simulation, we demonstrated the proposed approach with $3$ challenging rearrangement-based manipulation tasks, and compared its performance against $2$ baseline algorithms. In terms of efficiency and robustness, we showed that our approach is significantly faster and is able to complete tasks under perception uncertainties, local modeling errors, and nondeterministic physics in the real world.

In future work, we plan to incorporate motions that alternate the end-effector's poses across subsequent action segments, enabling the robot to quickly switch between different problem subspaces without being constrained to move continuously in $SE(2)$, so as to improve the efficiency of rearrangement actions. 
In addition, we plan to study the adaptation of the planning horizon in terms of other task-relevant factors, e.g., the number and distribution of concurrent contacts, in order to more tightly close the manipulation loop. Besides, we will also investigate the optimization of the local trajectories to reduce the execution time.

\bibliographystyle{IEEEtran}
\bibliography{refs}

\end{document}